\documentclass[letterpaper, 10 pt, conference]{ieeeconf}  

\IEEEoverridecommandlockouts                              

\usepackage{graphicx} 
\usepackage{amsmath} 
\usepackage{amssymb}  
\usepackage{bm}
\usepackage[ruled]{algorithm2e}
\usepackage{algpseudocode}
\usepackage{textcomp}
\usepackage{lipsum}
\usepackage{cuted}
\usepackage{gensymb}
\usepackage{multirow}
\usepackage{caption}
\usepackage[table]{xcolor}
\usepackage{array,colortbl}
\usepackage[noadjust]{cite}
\usepackage{hyperref}
\hypersetup{
    colorlinks=true,
    linkcolor=blue,
    filecolor=magenta,      
    urlcolor=cyan,
    }

\usepackage{xcolor}

\usepackage{array}
\newcolumntype{P}[1]{>{\centering\arraybackslash}p{#1}}

\title{\LARGE \bf
GelSight Svelte Hand: A Three-finger, Two-DoF, Tactile-rich, Low-cost Robot Hand for Dexterous Manipulation
}

\author{Jialiang (Alan) Zhao$^{1}$ and Edward H. Adelson$^{1}$
\thanks{$^{1}$Computer Science and Artificial Intelligence Lab, Massachusetts Institute of Technology
    {\tt\small \{alanzhao, adelson\}@csail.mit.edu}}
}

\begin{document}

\maketitle
\thispagestyle{empty}
\pagestyle{empty}


\begin{strip}
\centering
\includegraphics[width=\linewidth]{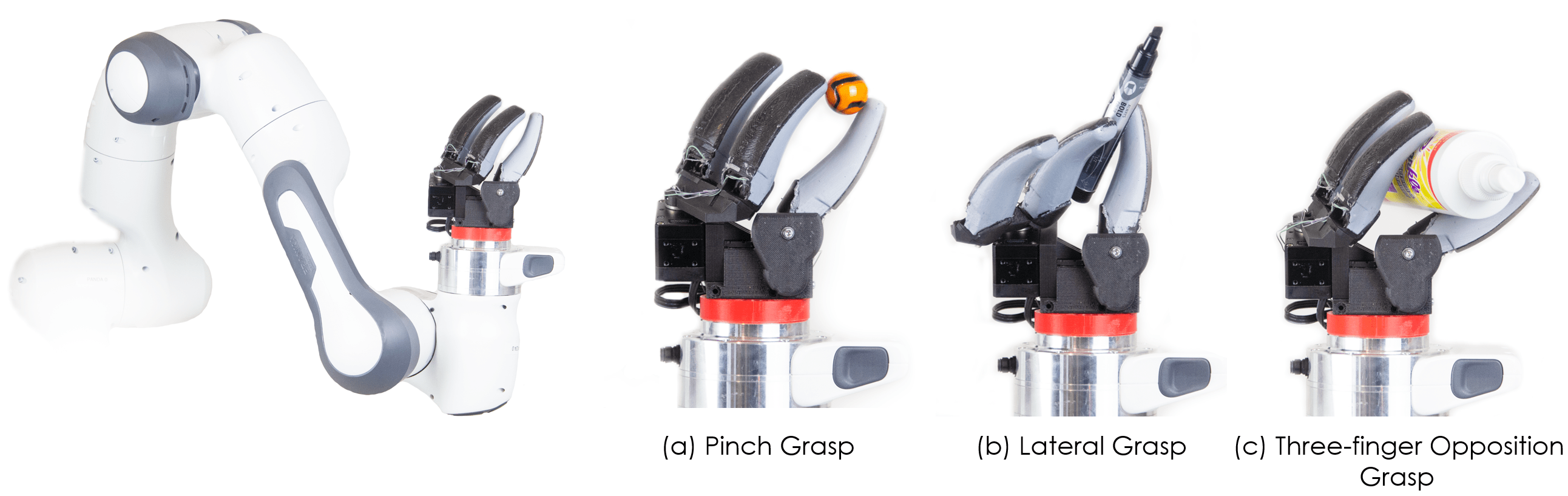}
\captionof{figure}{Different grasping modes of GelSight Svelte Hand and examples of their use cases. (a) Pinch grasp for small objects where high precision and low occlusion is needed. (b) Lateral grasp, an intermediate grasping mode where moderate precision and strength are both required. (c) Three-finger opposition grasp for larger or heavier objects where high strength is crucial.}
\label{fig:modes}
\end{strip}

\begin{abstract}
This paper presents GelSight Svelte Hand, a novel 3-finger 2-DoF tactile robotic hand that is capable of performing precision grasps, power grasps, and intermediate grasps.
Rich tactile signals are obtained from one camera on each finger, with an extended sensing area similar to the full length of a human finger.
Each finger of GelSight Svelte Hand is supported by a semi-rigid endoskeleton and covered with soft silicone materials, which provide both rigidity and compliance.
We describe the design, fabrication, functionalities, and tactile sensing capability of GelSight Svelte Hand in this paper.
More information is available on our website: \url{https://gelsight-svelte.alanz.info}.
\end{abstract}

\section{Introduction}

Human hand is dexterous thanks to the many grasping modes it is able to perform, the compliance and strength provided by the soft skin and the rigid skeleton, and the rich touch signals from all over the hand.
However, designing a robot hand with all those properties is challenging.
Generally, having more grasping modes means more individually actuated degrees of freedom are needed, which poses mechanical complexity and difficulty in control.
Soft fingers and hands actuated with pneumatic or hydraulic power are usually considered to have infinite degrees of freedom while also providing a soft contact with the in-hand object, however the highly compliant nature makes the state estimation and control problem challenging.
Most existing robot hands that are capable of high resolution tactile sensing are only sensorized at the fingertips, mainly due to space limitation and circuit complexity in order to cover a large sensing area.
It is thus desirable to have a soft, tactile robot hand that is able to perform multiple grasp modes without sacrificing manufacturability and cost. 



To this end, we introduce GelSight Svelte Hand with the following contributions:

\begin{itemize}
    \item A hand arrangement that achieves three commonly used grasping modes (pinch grasp, lateral grasp, and power grasp) using only two off-the-shelf servo motors.
    \item A hand design with three human finger-shaped tactile fingers, where each finger is constructed with a semi-flexible backbone and a soft silicone skin, which provides both rigidity and compliance.
    \item A hand that provides rich tactile sensing signals over a large area, while requiring only three low-cost cameras.
\end{itemize}


\begin{figure*}
\centering
\includegraphics[width=\linewidth]{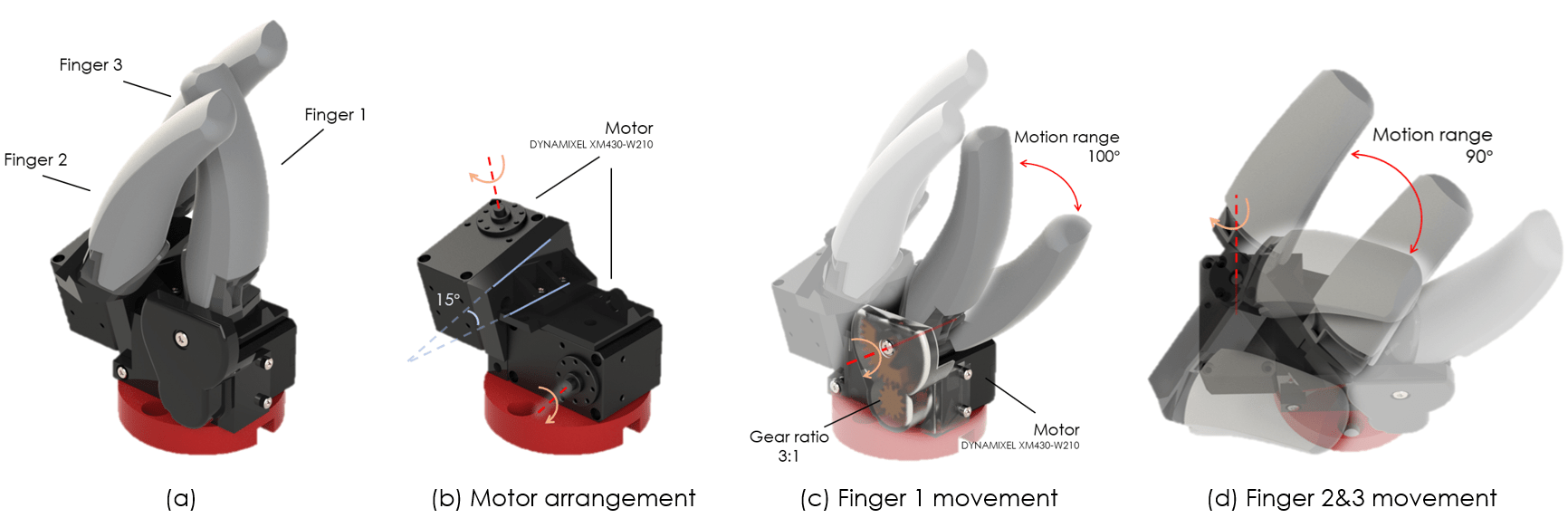}
\caption{Configuration and actuation of GelSight Svelte Hand. (a) Finger 1 functions similarly as a human thumb, where Finger 2 and Finger 3 together serve as a flipper. (b) One motor drives Finger 1, and the other motor drives Finger 2 and 3. The $15\degree$ offset between the two motors is critical to make the three grasping modes possible. (c) Finger 1 has a motion range of $100\degree$, and the gear ratio is 3:1 between the finger and the servo motor. (d) Finger 2 and 3 together are directly driven by the other motor, with a motion range of $90\degree$.}
\label{fig:actuation}
\end{figure*}

\section{Related Works}

Many anthropomorphic hands have been introduced aiming for a similar level of dexterity as that of human hands \cite{belter2016comparative, puhlmann2022rbo, mouri2011review, shadowhand}.
However, more degrees of freedom usually come with the cost of higher mechanical and control complexity.
Compared to conventional anthropomorphic hands, the design objective of GelSight Svelte Hand is not to replicate the structure of human hands or reproduce as many human grasping modes as possible, but to perform only the most commonly used grasping modes while keeping the mechanical complexity and the cost low.
The GRASP taxonomy classifies human grasps into three main categories: power, precision, and intermediate \cite{feix2015grasp}.
GelSight Svelte Hand is able to perform one grasp from each category: a three-finger opposition grasp for power grasps, a two-finger opposition grasp for precision grasps, and a two-finger lateral grasp for intermediate grasps.


Tactile sensing can be achieved by measuring pressure with different kinds of pressure sensors \cite{tomo2018new, shashank2009design, wen2008tuning}, or reconstructing deformation of a soft contact surface through a camera \cite{yuan2017gelsight, donlon2018gelslim, do2022densetact}.
Pressure sensor array-based tactile sensors are usually thinner and more modular, making them easier to be integrated on robots, such as the Honda 5-fingered hand \cite{dikhale2022visuotactile} that was built with 224 small pressure sensors that spread across the inside of the hand, and the iCub humanoid robot that is partially coverred with uSkin 3-axis pressure sensors \cite{tomo2018new}.
However, compared to camera-based tactile sensing solutions, pressure sensor arrays usually have much lower resolution.
The difficulty with integrating camera-based tactile sensors is that they usually require larger space due to illumination and optical constraints inside the sensor.
Therefore, most camera-based tactile sensors are used as fingertip sensors \cite{do2022densetact, lambeta2020digit}, leaving the majority surfaces of the hand unsensorized.


The closest robot hand design to our design is the GelSight EndoFlex hand \cite{liu2023gelsight}, a 3-fingered hand that is able to perform wrapping grasps and produce rich tactile signals.
Each of the three fingers on GelSight EndoFlex has two camera-based tactile sensing sections.
A flexible endoskeleton supports the finger.
When actuated, the endoskeleton is pulled by wires to close towards the center of the hand.
Compared to GelSight EndoFlex, GelSight Svelte Hand requires less motors (2 vs. 3), less cameras (3 vs. 6), and it is more versatile in terms of possible grasping modes (3 vs. 1).

\section{Design}
\label{sec:design}

\begin{figure}[h]
\centering
\includegraphics[width=\linewidth]{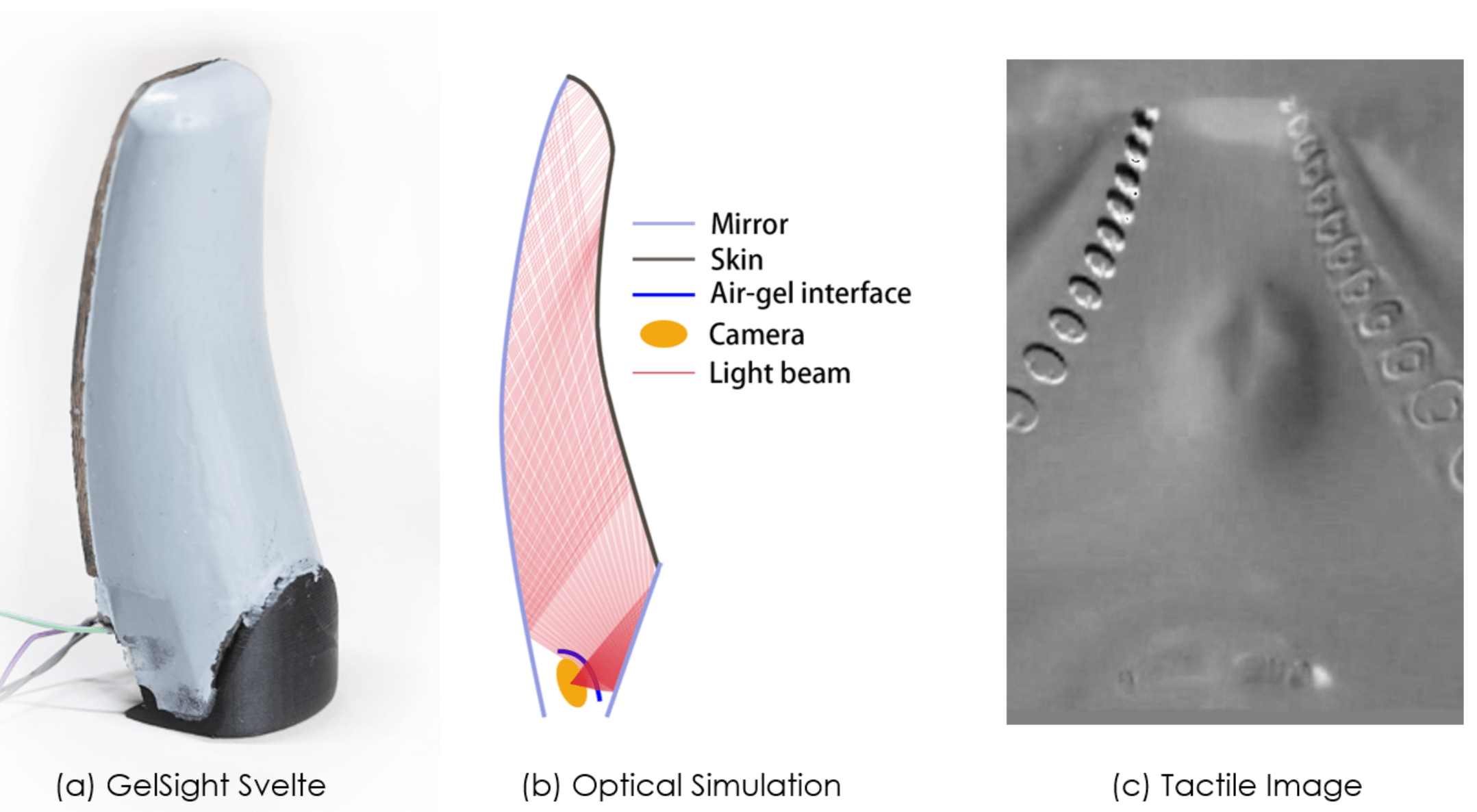}
\caption{(a) A GelSight Svelte tactile sensor. (b) The optical simulation of GelSight Svelte. The camera's FoV is able to cover the entire sensing skin with the help of two mirrors. (c) A tactile image obtained when pressing the head of a M8 screw against the middle section of the finger.}
\label{fig:finger}
\end{figure}

GelSight Svelte Hand is constructed with three GelSight Svelte tactile fingers, as shown in Fig. \ref{fig:actuation}(a). 
GelSight Svelte finger is a human finger-shaped, camera-based tactile sensor that has an elongated and curved sensing surface.
It utilizes curved mirrors inside the finger to achieve a relatively large sensing area using a single camera.
Its construction, optical simulation, and an example of tactile images is illustrated in Fig. \ref{fig:finger}.
We refer readers to \cite{zhao2023gelsight} for details about the design and analysis on this tactile finger.

Finger 1, or "the thumb", opposes the other two fingers and functions like a human thumb.
It is driven by a DYNAMIXEL XM430-W210 servo motor with a 3:1 gearing ratio, as shown in Fig. \ref{fig:actuation}(c).
Finger 2 and Finger 3, or "the flipper", rotates to form different grasp modes and they function like an index finger and a middle finger.
They are bridged together and directly driven by another DYNAMIXEL XM430-W210 servo motor, as shown in Fig. \ref{fig:actuation}(d).

As illustrated in Fig. \ref{fig:actuation}(b), the two motors are attached together with a $15\degree$ offset between them.
This offset makes the three grasping modes possible. 

\begin{figure}[h]
\centering
\includegraphics[width=\linewidth]{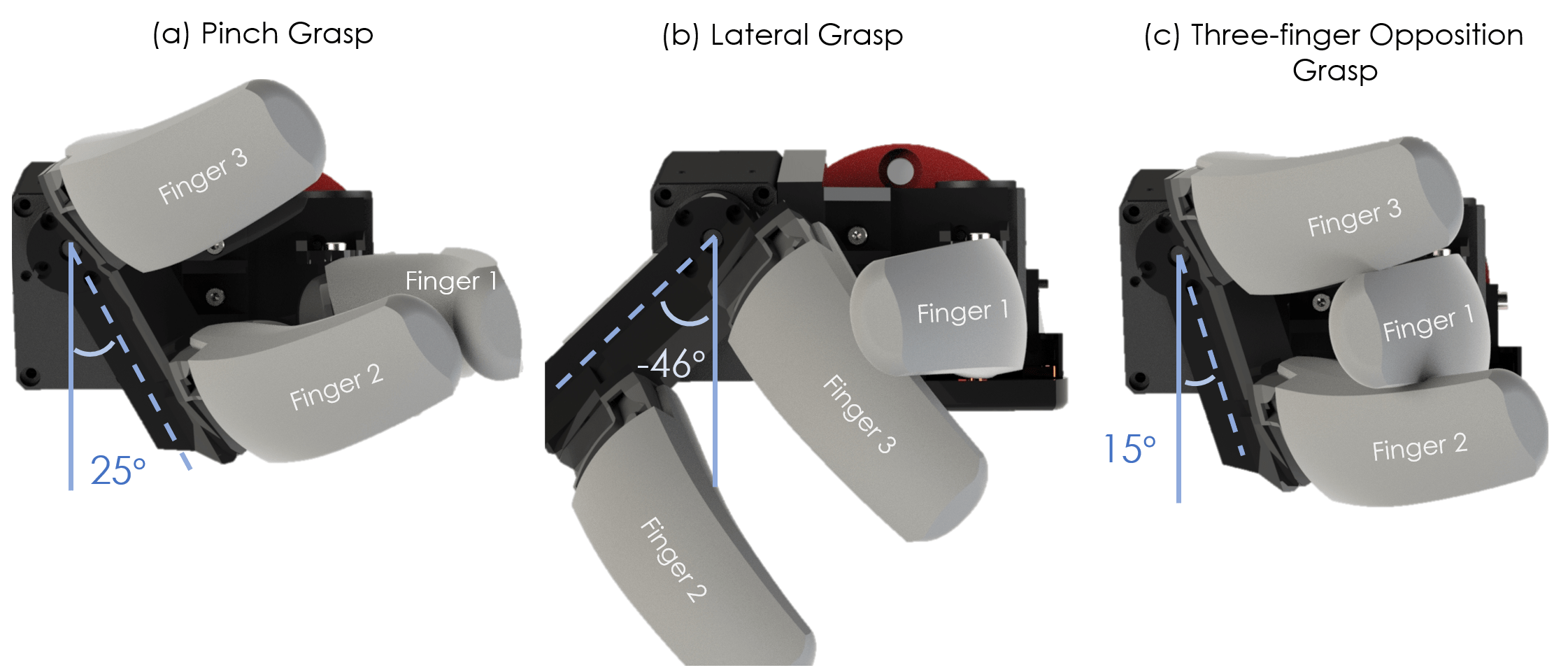}
\caption{All three grasping modes are activated in a similar fashion. The controller first opens Finger 1, then moves Finger 2 and 3 to a pre-specified position, at last closes Finger 1.}
\label{fig:motor_modes}
\end{figure}

The three grasping modes are illustrated in Fig. \ref{fig:motor_modes}.
All three grasping modes are performed in a similar fashion.
First, Finger 1 opens, with a position control mode, to a pre-determined open position.
Second, "the flipper" which contains Finger 2 and 3, moves to a specific location which is determined by the grasping mode, also with a position control mode.
The angles between the flipper bridge and the center axis of the flipper motor are $25\degree, -46\degree, 15\degree$ for pinch grasping, lateral grasping, three-finger opposition grasping, respectively.
At last, Finger 1 closes in a torque control mode.
Note that while Finger 1 closes, the flipper bridge is still engaged in the position control mode to avoid any unwanted movements due to load.

Mechanical specifications of GelSight Svelte Hand are listed in Tab. \ref{tab:specs}.

\begin{table}[h]
\caption {Specifications of GelSight Svelte Hand}
\label{tab:specs} 
\begin{center}
\begingroup

\setlength{\tabcolsep}{8pt} 
\renewcommand{\arraystretch}{1.5} 

\begin{tabular}{ |c|c|c| } 
\hline
\multirow{2}{10em}{Weight} & Each finger & $37g$ \\[0.2em] \cline{2-3}
& Entire hand & $361g$ \\[0.2em]
\hline
\multirow{2}{10em}{Maximum operational angles} & Finger 1 & $100\degree$ \\ \cline{2-3}
& Finger 2 \& 3 & $90\degree$ \\ 
\hline
\multirow{3}{10em}{Maximum force at fingertip} & All fingers (normal) & $2.5 N$ \\ \cline{2-3}
& Finger 1 (tangential) & $6.3 N$  \\  \cline{2-3}
& Finger 2\&3 (tangential) & $2.2 N$ \\ 
\hline
\multirow{2}{10em}{Gear ratio} & Finger 1 & $3:1$ \\  \cline{2-3}
& Finger 2 \& 3 & $1:1$ \\ 
\hline
\end{tabular}
\endgroup
\end{center}
\end{table}

\section{Application}

\begin{figure}[h]
\centering
\includegraphics[width=\linewidth]{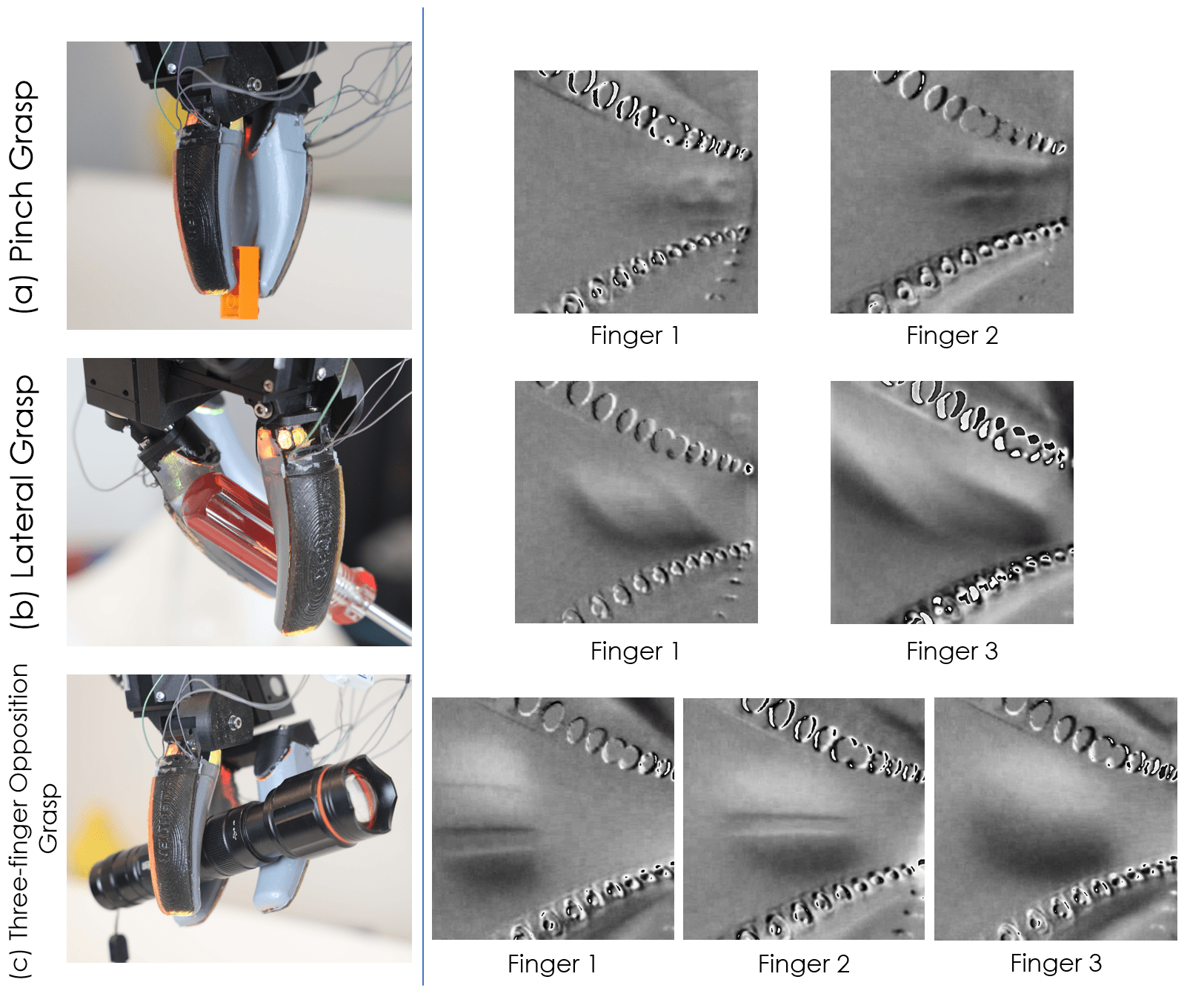}
\caption{The object holding experiment and the tactile images collected from relevant fingers. (a) A LEGO brick holding task with a pinch grasp. (b) A screw driver holding task with a lateral grasp. (c) A flash light holding task with a three-finger opposition grasp. Tactile images show textures of the contact between each finger and the object.}
\label{fig:sensing}
\end{figure}

To demonstrate the potential use cases for GelSight Svelte Hand, we conduct one object holding task with each of the three grasping modes, and show the tactile images collected from the relevant fingers.
Illustration and tactile images are shown in Fig. \ref{fig:sensing}.
More in-depth analysis of the tactile images in this task is given in \cite{zhao2023gelsight}.

\begin{figure}[h]
\centering
\includegraphics[width=0.6\linewidth]{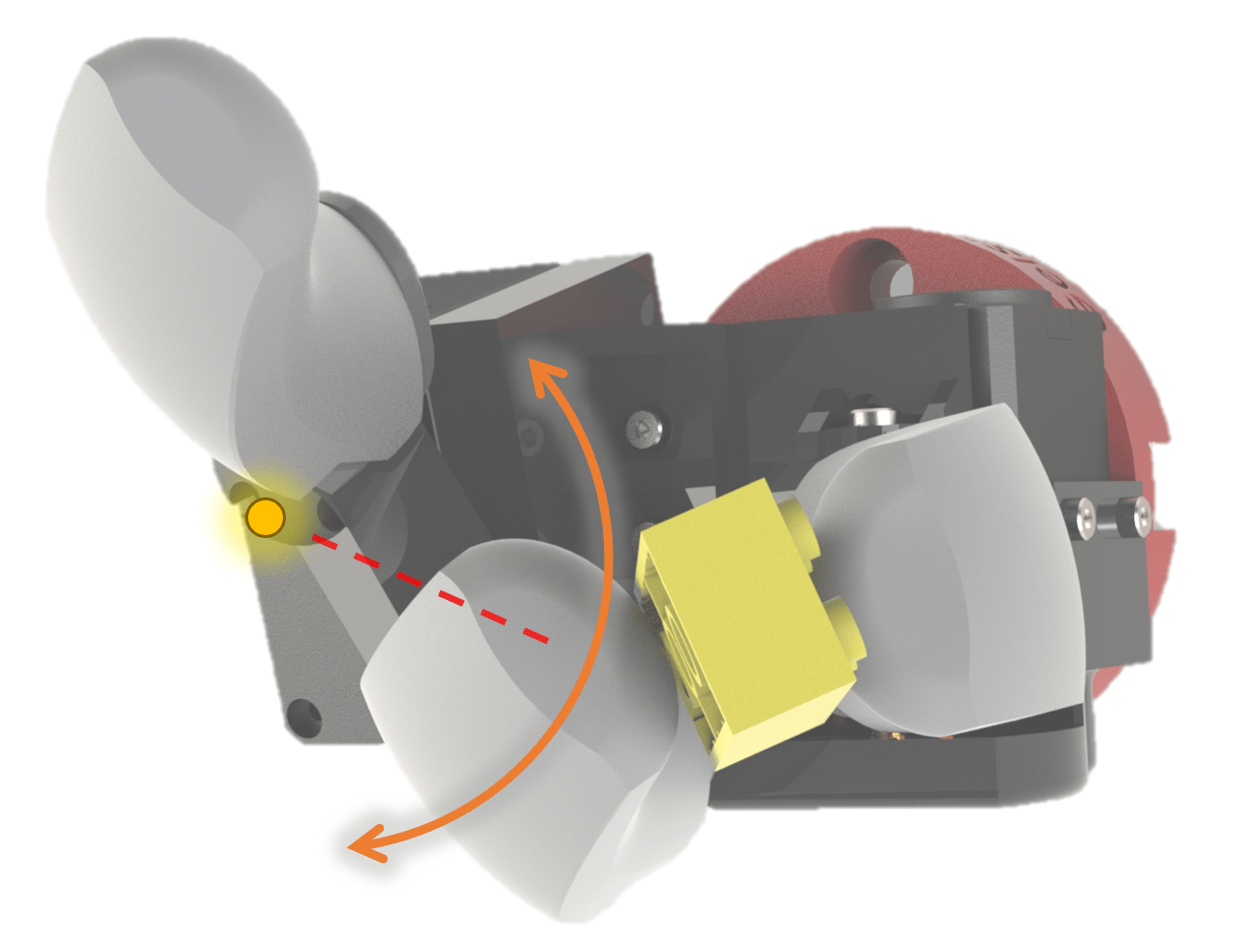}
\caption{A special twisting motion is possible when the hand is in a pinch grasping mode. The controller rotates Finger 2 and 3 within a small range, while still applying force on Finger 1 to achieve this motion.}
\label{fig:twisting}
\end{figure}

\textbf{Pinch grasping} is demonstrated with a LEGO brick grasping task, which requires high precision and low occlusion.
Only Finger 1 and Finger 2 are utilized in this task, and the contact is constrained at the fingertips.
GelSight Svelte Hand is able to grasp the 2x4 LEGO brick in a similar way as human hands do, and the tactile sensing signals provide unique insight into the state of the LEGO brick as shown in Fig. \ref{fig:sensing}(a).
An additional twisting motion can be made possible by actuating Finger 2 to move in a small range, as illustrated in Fig. \ref{fig:twisting}.
In a pinch grasping configuration, the largest opening between the tips of Finger 1 and Finger 2 is $63mm$.

\textbf{Lateral grasp} is one example of \textit{intermediate grasps}, which lies between "power grip" and "precision handling" \cite{feix2015grasp}.
One typical intermediate grasp of human hands uses the fingertip of the thumb and the side surface of the index finger, such as the way one holds a key or a pencil.
GelSight Svelte Hand recreates this grasping mode by opposing the fingertip of Finger 1 and the side surface of Finger 3, as shown in Fig. \ref{fig:sensing}(b).
A screw-driver holding task is performed to demonstrate this grasping mode.
Thanks to the larger sensing surface and the round shape of GelSight Svelte, rich contact information can be obtained on both fingers.
In this configuration, the largest opening between the middle section of Finger 1 and the side of Finger 3 is $80mm$.

\textbf{Three-finger opposition grasp} produces the most stability and strength. 
It is demonstrated with a flash light holding task in Fig. \ref{fig:sensing}(c).
All three fingers are utilized to provide the maximum stability and strength.
The maximum opening is $72mm$.

\section{CONCLUSIONS}

In this paper we present GelSight Svelte Hand, a 3-finger, 2-DoF robotic design that allows precision, intermediate, and power grasps while providing rich tactile signals during manipulation.
Mechanical configurations and specification are discussed, and one demonstration task for each grasping mode is given alongside the tactile images.
To the authors' best knowledge, GelSight Svelte Hand is the first design that utilizes such a mechanical configuration for versatile operations and tactile signal acquisition.








\section*{ACKNOWLEDGMENT}

Toyota Research Institute provided funds to support this work.


\bibliographystyle{IEEEtran}
\bibliography{refs}

\end{document}